# TRANSFER LEARNING FROM IMAGENET FOR MEG-BASED DECODING OF IMAGINED SPEECH

*Soufiane Jhilal[1], Stéphanie Martin[2], Anne-Lise Giraud[1]*

[1] Université Paris Cité, Institut Pasteur, AP-HP, INSERM, CNRS, Fondation Pour l'Audition, Institut de l'Audition, IHU reConnect, Paris, France
[2] Department of Basic Neurosciences, Faculty of Medicine, University of Geneva, Geneva, Switzerland

## ABSTRACT

Non-invasive decoding of imagined speech remains challenging due to weak, distributed signals and limited labeled data. Our paper introduces an image-based approach that transforms magnetoencephalography (MEG) signals into time-frequency representations compatible with pretrained vision models. MEG data from 21 participants performing imagined speech tasks were projected into three spatial scalogram mixtures via a learnable sensor-space convolution, producing compact image-like inputs for ImageNet-pretrained vision architectures. These models outperformed classical and non-pretrained models, achieving up to 90.4% balanced accuracy for imagery vs. silence, 81.0% vs. silent reading, and 60.6% for vowel decoding. Cross-subject evaluation confirmed that pretrained models capture shared neural representations, and temporal analyses localized discriminative information to imagery-locked intervals. These findings show that pretrained vision models applied to image-based MEG representations can effectively capture the structure of imagined speech in non-invasive neural signals.

## 1. INTRODUCTION

Decoding cognitive and perceptual states from non-invasive neural recordings constitutes a key challenge for scalable neurotechnology. Progress in this area increasingly depends on developing representations that can capture the spatial, spectral, and temporal structure of neural signals in a form suitable for modern machine learning methods. Decoding speech-related brain activity is of particular interest for its potential in communication restoration. Intracranial brain-computer interfaces (BCIs) have recently demonstrated impressive results [1], by decoding attempted [2] and inner speech [3], and even reconstructing intelligible audio [4]. However, their reliance on invasive recordings limits broader applicability. This motivates non-invasive approaches that can extract meaningful representations from neural signals while maintaining generalization across individuals.

Imagined speech poses a particularly challenging problem as it produces no measurable output, yet remains the only option for those unable to speak. The absence of external feedback limits data availability and supervised training, making transfer learning from large-scale domains especially valuable. Leveraging computer vision models trained on vast image datasets could offer a principled way to exploit transferable feature hierarchies for decoding imagined speech from non-invasive recordings.

While electroencephalography (EEG) has been widely explored for speech decoding [5], [6], [7], its limited bandwidth and spatial resolution restrict feature diversity. Magnetoencephalography (MEG) provides higher bandwidth and better spatial precision, making it an ideal testbed for developing new representation-learning methods. As recent works have shown promising results in decoding perceived language [8], typed characters [9], and imagined sentences [10] from MEG, we propose an image-based time-frequency (TF) representation that transforms MEG activity into a form amenable to transfer learning with pretrained vision models.

In summary, this work (i) introduces a compact image-based MEG representation that learns three sensor-space maps from time–frequency scalograms, (ii) systematically evaluates ImageNet-pretrained models against classical and neural baselines on three imagined speech decoding tasks, and (iii) provides an ablation quantifying the roles of transfer learning, sensor mixing, and fine-tuning strategy in non-invasive speech decoding.

## 2. METHOD

### 2.1. Dataset and tasks

Twenty-six neurotypical native French speakers performed an imagined speech task (Fig. 1); five were excluded due to recording or task-related issues, leaving 21 participants. The experiment consisted of 3 blocks of 34 sentences (6 to 11 syllables long). Each syllable was drawn from a closed set of 18 items combining one of five consonants (/l/, /m/, /r/, /s/, /t/) or none with one of three vowels (/a/, /e/, /i/). Each trial comprised silent reading (SR) followed by imagined speech production (ISP) of a sentence at 400 ms per syllable. MEG was recorded at 1 kHz with 248 magnetometers.

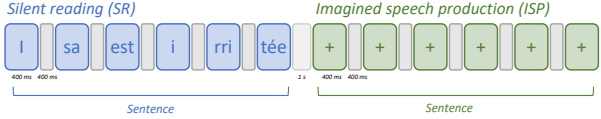

**Fig. 1.** Imagined speech paradigm. Each trial comprised silent reading (SR) followed by imagined speech production (ISP) of the same sentence at 400 ms per syllable. MEG was recorded across three blocks of 34 sentences each.

Three decoding tasks were analyzed:
- **ISP vs Silence:** discriminating ISP from silence segments.
- **ISP vs SR:** differentiating ISP from SR segments.
- **Vowel Decoding:** classifying /a/, /e/, /i/ during ISP.

For vowel decoding, syllables were grouped by vowel nucleus (/a/, /e/, /i/), reflecting the dominant vocalic content (the most acoustically salient feature in speech) and addressing limited trial counts per syllable class.

### 2.2. Preprocessing and epoch extraction

MEG signals were notch-filtered at 50, 100 and 150 Hz, band-passed (0.5-150 Hz), downsampled to 500 Hz. Artifacts were removed using independent component analysis (ICA). Bad segments were identified using AutoReject [11]. The data were then epoched relative to cue onset into three windows: pre-cue (-300 to 0 ms), post-cue (0 to 500 ms), and full window (-300 to 500 ms). Decoding was run separately on each window to compare anticipatory, evoked, and combined representations. Silence epochs were drawn from inter-trial rest periods, not from inter-syllable gaps. Across all subjects and after preprocessing, we were left with around 12~13 thousand epochs per class for ISP vs Silence and ISP vs SR, and approximately 4 thousand epochs per vowel.

### 2.3. Time-frequency representation

Morlet continuous wavelet transform (CWT) was applied to each channel independently, using 96 logarithmically spaced frequencies (1-150 Hz). This produced complex coefficients $W(f,t)$ for each frequency $f$ and time point $t$ within the epoch window. The magnitude $|W(f,t)|$ yielded time-frequency scalograms with 96 frequency bins and variable time points depending on epoch duration. Each scalogram was z-scored relative to baseline to normalize power fluctuations.

### 2.4. Conversion to image-like representations

For each epoch, the 248-channel MEG scalograms were concatenated into a tensor of shape (248, F, T), where F=96 frequency bins and T depends on the epoch length (150, 250 or 400 time points for pre-, post-, or full windows). To obtain a 3-channel representation compatible with pretrained vision models, we applied a learnable 1×1 convolutional projection across the sensor dimension. This sensor-space mixing layer, inspired by the channel-mixing operation in ConvMixer networks [12], learns distributed neural patterns while reducing dimensionality. The result was a tensor of shape (3, F, T), analogous to RGB image channels where the axes encode frequency, time, and learned spatial mixtures. Each tensor was resized using bilinear interpolation to match the 224×224×3 input expected by ImageNet [13] pretrained architectures. The full pipeline is illustrated in Fig. 2. We also tested a source-space variant that yielded similar decoding performance. We thus only report the sensor-space pipeline for brevity.

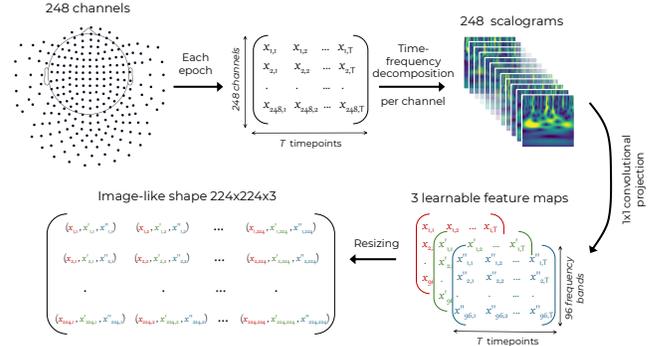

**Fig. 2.** Data processing pipeline. Time–frequency scalograms from all sensors were concatenated, projected into three spatial maps via a 1×1 convolution, and resized to 224×224×3 for input to pretrained vision models.

### 2.5. Classification Models

**Pretrained Vision Models:** ResNet-18 [14] and ViT-Tiny [15] pretrained on ImageNet were used with the TF-image epochs described above. Final layers were replaced with task-specific heads (sigmoid for binary tasks; 3-way softmax for vowel decoding). Early and mid-level layers were kept frozen, while the final convolutional or transformer block, the new classification layers, and the learnable 1×1 convolutional projection were trained from scratch. Differential learning rates were used: $1\times10^{-3}$ for the classification heads and the convolutional layers, and $1\times10^{-4}$ for the unfrozen layers.

**Non-pretrained Neural Baselines**: A shallow convolutional neural network (CNN) (4 layers: 32→64→128 filters, 3×3 kernels, dropout 0.3) was trained on the same TF-image inputs as the pretrained vision models. EEGNet (compact version from [16]) was trained using raw time-domain sensor data (248 channels × T samples per epoch). Inputs were z-scored per channel.

**Classical Baselines**: Linear Discriminant Analysis (LDA) was trained on flattened TF-image tensors. For a stronger MEG/EEG baseline, we used a Riemannian Tangent-Space classifier: for each epoch, covariance matrices (248×248) were computed, projected to tangent space at the Riemannian mean, then the upper-triangular vectorization was classified with L2-regularized logistic regression.

**Training**: All neural models used weighted cross-entropy loss, AdamW optimizer, and cosine annealing over 30 epochs. Data augmentation included temporal shifts (±50~100 ms) within the same condition window, frequency masking, and ±5% amplitude jitter. Batch size was 32 with gradient clipping (max-norm 1.0).

**Evaluation**: Two protocols were used: Subject-Agnostic Pooled (SAP) with stratified 10-fold cross-validation grouped by trial to avoid within-trial leakage, and Leave-One-Subject-Out (LOSO) training on N-1 participants (21 folds), with strict separation between training and test splits. Performance was quantified using balanced accuracy to mitigate class imbalance. Chance level was 50% for binary tasks, and 33% for 3-class vowel decoding. Metrics were reported with 95% bootstrap confidence intervals (CIs). Statistical significance was assessed using Wilcoxon signed-rank tests for pairwise model comparisons (two-sided) and permutation tests (10,000 iterations) for comparisons against chance. All p-values were corrected for multiple comparisons using the Holm-Bonferroni method.

## 3. RESULTS

### 3.1. Imagined Speech Production vs Silence

As anticipated, decoding imagined speech from silence yielded the highest accuracies across all tasks (Table 1, Fig. 3). Performance rose from pre-cue to post-cue and full windows, indicating that most discriminative information was present during imagery. ResNet-18 achieved the top scores (90.4% SAP, 77.5% LOSO) in the full window, followed closely by ViT-Tiny. Post-cue and full-window performance matched closely, suggesting stimulus-locked neural signatures.

**Table 1.** ISP vs Silence decoding performance. Balanced accuracy (mean ± 95% CI) for each model, time window, and evaluation protocol (SAP = Subject-Agnostic Pooled; LOSO = Leave-One-Subject-Out). Chance level is 50%. Highest scores are in bold.

| | Pre-cue | | Post-cue | | Full window | |
|---|---|---|---|---|---|---|
| | SAP | LOSO | SAP | LOSO | SAP | LOSO |
| LDA | 56.6 ± 2.4 | 53.1 ± 0.8 | 66.3 ± 3.3 | 61.2 ± 1.2 | 66.4 ± 2.3 | 62.5 ± 0.9 |
| EEGNet | 61.0 ± 4.4 | 56.6 ± 0.8 | 74.4 ± 2.1 | 69.2 ± 0.7 | 77.1 ± 1.0 | 69.9 ± 0.7 |
| Shallow CNN | 64.6 ± 2.8 | 56.7 ± 0.8 | 79.7 ± 1.7 | 69.6 ± 1.0 | 81.4 ± 2.7 | 72.0 ± 1.0 |
| Riemannian | 63.5 ± 3.6 | 55.8 ± 0.9 | 80.1 ± 2.1 | 70.4 ± 1.2 | 81.7 ± 2.5 | 71.8 ± 1.5 |
| ResNet-18 | **72.9 ± 2.5** | 60.8 ± 0.8 | **90.2 ± 2.7** | **77.0 ± 1.4** | **90.4 ± 2.5** | **77.5 ± 1.3** |
| ViT-Tiny | 70.3 ± 2.4 | **61.2 ± 0.8** | 86.4 ± 2.4 | 75.7 ± 1.1 | 88.8 ± 2.9 | 77.0 ± 1.4 |

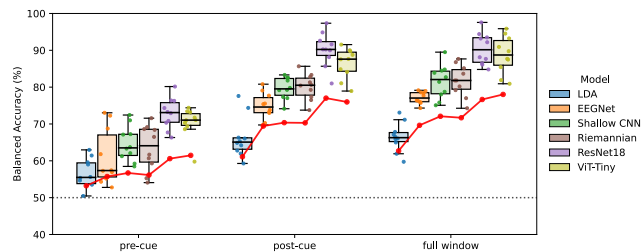

**Fig. 3.** ISP vs Silence decoding accuracies. Boxplots show SAP balanced-accuracy distributions across participants for each model and time window (pre-cue, post-cue, full). Red lines connect corresponding LOSO medians, and the dotted line indicates the 50% chance level.

Pretrained vision models significantly outperformed baseline models (p < 0.05). ResNet-18 outscored the shallow CNN (p = 0.003) and EEGNet (p < 0.001), confirming pretraining benefits beyond architecture. LOSO accuracies remained significantly above chance for pretrained models (p < 0.001).

### 3.2. Imagined Speech Production vs Silent Reading

Distinguishing imagined from silently read sentences proved more challenging (Table 2, Fig. 4). While accuracies were lower overall, pretrained architectures still yielded clear gains over baseline models (p < 0.05), with ResNet-18 again reaching the best accuracy (81.0% SAP, 72.2% LOSO).

**Table 2.** ISP vs SR decoding performance. Balanced accuracy (mean ± 95% CI) for each model, time window, and evaluation protocol (SAP = Subject-Agnostic Pooled; LOSO = Leave-One-Subject-Out). Chance level is 50%. Highest scores are in bold.

| | Pre-cue | | Post-cue | | Full window | |
|---|---|---|---|---|---|---|
| | SAP | LOSO | SAP | LOSO | SAP | LOSO |
| LDA | 51.7 ± 3.6 | 50.5 ± 0.7 | 61.6 ± 3.2 | 57.7 ± 1.1 | 64.6 ± 2.7 | 60.5 ± 1.2 |
| EEGNet | 53.2 ± 3.0 | 52.5 ± 0.8 | 70.5 ± 2.7 | 65.4 ± 1.2 | 72.3 ± 2.5 | 67.1 ± 1.5 |
| Shallow CNN | 54.9 ± 2.9 | 52.7 ± 0.6 | 72.9 ± 1.9 | 65.6 ± 1.2 | 74.1 ± 2.0 | 67.8 ± 1.4 |
| Riemannian | 54.9 ± 3.8 | 52.4 ± 0.9 | 72.1 ± 2.3 | 66.6 ± 1.5 | 74.6 ± 2.7 | 66.7 ± 1.2 |
| ResNet-18 | **61.9 ± 2.2** | **57.0 ± 0.9** | **79.1 ± 2.4** | 70.0 ± 1.4 | **81.0 ± 2.0** | **72.2 ± 1.5** |
| ViT-Tiny | 59.6 ± 1.5 | 56.3 ± 0.9 | 76.3 ± 3.4 | **70.3 ± 1.1** | 79.7 ± 2.7 | 71.4 ± 1.2 |

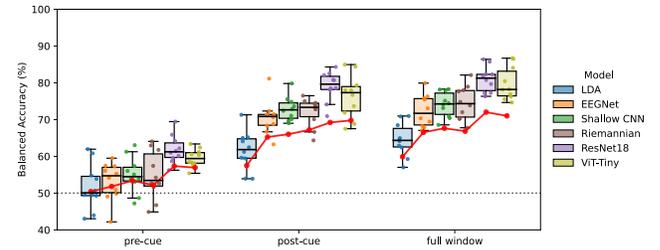

**Fig. 4.** ISP vs SR decoding accuracies. Boxplots show SAP balanced-accuracy distributions across participants for each model and time window (pre-cue, post-cue, full). Red lines connect corresponding LOSO medians, and the dotted line indicates the 50% chance level.

The performance drop for ISP vs. SR, compared to ISP vs Silence (81.0% vs 90.4%, p < 0.001) reflects the increased difficulty of discriminating between two covert speech modes that share phonological processing pathways.

### 3.3. Vowel decoding

All models exceeded chance (p < 0.001) for post-cue and full window, confirming discriminable phonological features in imagined speech (Table 3, Fig. 5). ResNet-18 and ViT-Tiny achieved the highest performance (≈ 60% SAP, 51% LOSO), with a significant margin over baseline models (p < 0.05). The performance gap between binary (90.4%) and three-class (60.6%) tasks was significant (p < 0.001), highlighting the challenge of fine-grained phonemic discrimination during speech imagery from non-invasive recordings.

**Table 3.** 3-class vowel (/a/, /e/, /i/) decoding performance. Balanced accuracy (mean ± 95% CI) for each model, time window, and evaluation protocol (SAP = Subject-Agnostic Pooled; LOSO = Leave-One-Subject-Out). Chance level is 33%. Highest scores are in bold.

| | Pre-cue | | Post-cue | | Full window | |
|---|---|---|---|---|---|---|
| | SAP | LOSO | SAP | LOSO | SAP | LOSO |
| LDA | 33.8 ± 1.3 | 32.8 ± 0.6 | 43.1 ± 2.0 | 39.9 ± 0.8 | 43.2 ± 1.4 | 40.7 ± 0.5 |
| EEGNet | 34.4 ± 1.7 | 33.3 ± 0.6 | 48.5 ± 1.3 | 44.0 ± 0.4 | 49.9 ± 0.6 | 44.4 ± 0.4 |
| Shallow CNN | 36.0 ± 1.5 | 34.1 ± 0.6 | 53.4 ± 1.0 | 46.1 ± 0.6 | 54.5 ± 1.7 | 47.6 ± 0.6 |
| Riemannian | 35.7 ± 1.9 | 34.2 ± 0.8 | 53.6 ± 1.3 | 46.6 ± 0.7 | 54.6 ± 1.6 | 47.5 ± 0.9 |
| ResNet-18 | 35.9 ± 1.5 | 34.8 ± 0.7 | **60.5 ± 1.7** | **51.4 ± 0.9** | **60.6 ± 1.5** | **51.7 ± 0.8** |
| ViT-Tiny | **36.7 ± 1.5** | **35.0 ± 0.7** | 58.6 ± 1.5 | 50.1 ± 0.7 | 60.1 ± 1.8 | 51.0 ± 0.9 |

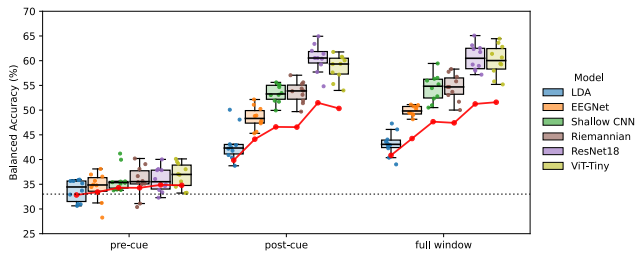

**Fig. 5.** 3-class vowel (/a/, /e/, /i/) decoding accuracies. Boxplots show SAP balanced-accuracy distributions across participants for each model and time window (pre-cue, post-cue, full). Red lines connect corresponding LOSO medians, and the dotted line indicates the 33% chance level.

### 3.4. Ablation analysis

We conducted an ablation on the best-performing model (ResNet-18, full window -300 to 500 ms; Table 4) to identify the architectural components driving performance gains. We systematically removed or altered key design elements; ImageNet pretraining, the learnable convolutional sensor projection, and the partial fine-tuning (FT) strategy, to quantify their individual contributions.

**Table 4.** Ablation analysis (ResNet-18, full window -300 to 500 ms). Balanced accuracy (mean ± 95% CI) for three decoding tasks under SAP (10-fold, subject-agnostic) and LOSO (21 folds). Baseline uses Convolutional projection + ImageNet pretraining with partial fine-tuning. Variants swap Convolutional projection for PCA-3, remove pretraining (RandomInit), and/or fully fine-tune (full FT). Highest scores are in bold.

| ISP vs Silence | | |
|---|---|---|
| **Variants (ResNet-18, full window)** | **SAP** | **LOSO** |
| **Baseline: Convolutional projection + Pretrained + partial FT** | **90.4 ± 2.5** | **77.5 ± 1.3** |
| Convolutional projection + Pretrained + full FT | 87.4 ± 2.7 | 74.8 ± 1.7 |
| PCA-3 + Pretrained + partial FT | 86.5 ± 2.6 | 73.5 ± 1.6 |
| PCA-3 + Pretrained + full FT | 85.1 ± 2.8 | 72.2 ± 1.8 |
| Convolutional projection + RandomInit + partial FT | 81.6 ± 2.8 | 70.8 ± 1.7 |
| Convolutional projection + RandomInit + full FT | 80.2 ± 2.9 | 69.1 ± 1.9 |
| PCA-3 + RandomInit + partial FT | 78.9 ± 2.9 | 67.3 ± 1.9 |
| PCA-3 + RandomInit + full FT | 77.6 ± 3.0 | 66.0 ± 2.0 |
| **ISP vs SR** | | |
| **Baseline: Convolutional projection + Pretrained + partial FT** | **81.0 ± 2.0** | **72.2 ± 1.5** |
| Convolutional projection + Pretrained + full FT | 77.8 ± 2.3 | 69.3 ± 1.8 |
| PCA-3 + Pretrained + partial FT | 76.6 ± 2.3 | 68.0 ± 1.7 |
| PCA-3 + Pretrained + full FT | 75.1 ± 2.5 | 66.7 ± 1.9 |
| Convolutional projection + RandomInit + partial FT | 72.1 ± 2.6 | 64.9 ± 1.8 |
| Convolutional projection + RandomInit + full FT | 70.8 ± 2.7 | 63.2 ± 2.0 |
| PCA-3 + RandomInit + partial FT | 69.4 ± 2.8 | 62.0 ± 2.0 |
| PCA-3 + RandomInit + full FT | 68.0 ± 2.9 | 60.7 ± 2.1 |
| **Vowel (/a/,/e/,/i/) decoding** | | |
| **Baseline: Convolutional projection + Pretrained + partial FT** | **60.6 ± 1.5** | **51.7 ± 0.8** |
| Convolutional projection + Pretrained + full FT | 58.0 ± 1.7 | 48.9 ± 1.0 |
| PCA-3 + Pretrained + partial FT | 57.1 ± 1.7 | 48.2 ± 1.0 |
| PCA-3 + Pretrained + full FT | 55.8 ± 1.8 | 47.1 ± 1.1 |
| Convolutional projection + RandomInit + partial FT | 53.3 ± 1.8 | 46.0 ± 1.1 |
| Convolutional projection + RandomInit + full FT | 51.9 ± 1.9 | 44.8 ± 1.2 |
| PCA-3 + RandomInit + partial FT | 50.6 ± 1.9 | 43.6 ± 1.2 |
| PCA-3 + RandomInit + full FT | 49.2 ± 2.0 | 42.3 ± 1.3 |

Removing pretraining caused the largest performance drop, confirming the benefit of transfer learning. Replacing convolutional with a fixed PCA-3 projection (first three principal components) also reduced accuracy, showing the value of learned spatial mixing. Fully fine-tuning all layers slightly lowered SAP and further reduced LOSO, indicating mild overfitting. Overall, the Convolutional projection + Pretrained + partial FT architecture outperformed all variants (Wilcoxon signed-rank test, $p < 0.05$), achieving the best decoding and cross-subject generalization.

### 4. DISCUSSION

Pretrained vision backbones outperformed classical baselines and a shallow CNN, showing that image-derived inductive biases effectively transfer to MEG TF representations. The proposed image-based formulation, combining a learnable sensor-space projection with pretrained models, enabled reliable decoding of imagined speech across all tasks. Together with the ablation analysis, which revealed the largest performance drops when removing ImageNet pretraining or the learnable projection, these findings demonstrate that convolutional and attention-based architectures trained on natural images can be repurposed to capture spatial-spectral-temporal structure in MEG data.

Decoding accuracy increased from pre-cue to post-cue and full windows, showing that most discriminative information is time-locked to imagery. Pre-cue performance was above chance for coarse ISP detection but near chance for vowel identity, indicating that anticipatory signals contain predictive information, whereas fine-grained phonological content emerges primarily during active imagery. The lower accuracy for ISP vs SR, compared to ISP vs. Silence, reflects overlapping processing between silent reading and imagery, with residual differences likely driven by articulatory simulation and sustained visual or phonological activation.

Cross-subject evaluation (LOSO) highlighted the challenge of generalization in neuroimaging, with lower performance than pooled evaluation (SAP). Yet pretrained models remained clearly above chance (up to 77.5% ISP vs Silence, 72.2% ISP vs SR, 51.7% vowels), suggesting shared spatial-spectral structure across participants captured by the three-map TF representation. ISP vs Silence decoding may nevertheless be influenced by broader differences in task engagement, attention, or residual sensorimotor activity, and current vowel accuracies remain below what would be required for practical speech BCIs. Future work will extend to larger phonemic sets and adaptive timing, incorporate sensor-weight topographies for interpretability, and explore domain adaptation and self-supervised pretraining on large MEG data to further enhance subject-independent decoding.

### 5. CONCLUSION

We show that pretrained vision architectures, combined with a learnable image-based MEG representation, effectively capture non-invasive neural dynamics underlying imagined speech. Temporal analyses revealed that discriminative information is concentrated in imagery-locked intervals, emphasizing the role of evoked frequency-time structure. These results highlight an imaging-driven strategy for neural decoding, where transforming MEG into compact TF images enables transfer learning and generalization across individuals, offering a promising methodological foundation for future non-invasive speech decoding studies.

## 6. COMPLIANCE WITH ETHICAL STANDARDS

## 7. ACKNOWLEDGMENTS

This work was supported by a grant from Fondation Pour l'Audition (FPA) to AL.G. (FPA IDA11) and benefited from a French government grant managed by the Agence Nationale de la Recherche under the France 2030 programme, reference ANR-23-IAHU-0003. We thank Ciprian Bangu for help at the initial stage of this work, and Emmanuel Mandonnet for his helpful feedback on the manuscript.